\documentclass[letterpaper,twocolumn,fleqn]{article} 

\usepackage{ist}
\usepackage{hyperref}
\hypersetup{
    colorlinks=true,
    linkcolor=blue,
    filecolor=magenta,      
    urlcolor=cyan,
    pdftitle={Overleaf Example},
    pdfpagemode=FullScreen,
    }
\usepackage{textcomp}
\usepackage{upgreek}
\usepackage{mathrsfs}
\usepackage{array}
\usepackage{graphicx}
\newcommand\smallbullet{%
    \raisebox{-0.25ex}{\scalebox{1.2}{$\cdot$}}%
}
\usepackage[dvipsnames]{xcolor}
\usepackage{caption}
\usepackage{adjustbox}

\title{Evaluating the Efficacy of Skincare Product: A Realistic Short-Term Facial Pore Simulation}

\author{Ling Li\textsuperscript{1}, Bandara Dissanayake\textsuperscript{2}, Tatsuya Omotezako\textsuperscript{2}, Yunjie Zhong\textsuperscript{1}, Qing Zhang\textsuperscript{3}, Rizhao Cai\textsuperscript{1}, Qian Zheng\textsuperscript{4}, Dennis Sng\textsuperscript{1}, Weisi Lin\textsuperscript{1}, Yufei Wang\textsuperscript{5}, Alex C Kot\textsuperscript{1};
\\
\textsuperscript{1} Nanyang Technological University, Singapore; \textsuperscript{2} The Procter \& Gamble Company; \textsuperscript{3} East China Normal University, China;
\\
\textsuperscript{4} Zhejiang University, China; \textsuperscript{5} China-Singapore International Joint Research Institute, China
\\
}

\date{} 

\begin{document} 

\maketitle 

\thispagestyle{empty} 



\begin{abstract}
Simulating the effects of skincare products on face is a potential new way to communicate the efficacy of skincare products in skin diagnostics and product recommendations. Furthermore, such simulations enable one to anticipate his/her skin conditions and better manage skin health. However, there is a lack of effective simulations today. In this paper, we propose the first simulation model to reveal facial pore changes after using skincare products. Our simulation pipeline consists of 2 steps: training data establishment and facial pore simulation. To establish training data, we collect face images with various pore quality indexes from short-term (8-weeks) clinical studies. People often experience significant skin fluctuations (due to natural rhythms, external stressors, etc.,), which introduces large perturbations in clinical data. To address this problem, we propose a sliding window mechanism to clean data and select representative index(es) to represent facial pore changes. Facial pore simulation stage consists of 3 modules: UNet-based segmentation module to localize facial pores; regression module to predict time-dependent warping hyperparameters; and deformation module, taking warping hyperparameters and pore segmentation labels as inputs, to precisely deform pores accordingly. The proposed simulation is able to render realistic facial pore changes. And this work will pave the way for future research in facial skin simulation and skincare product developments. 
\end{abstract}

\section{Introduction}
\label{sec:intro}
Consumers prefer smooth and flawless skin that makes them look youthful and healthy. Skin texture plays a key role in the perception of human facial beauty \cite{FinkHuman2001, HumphreyDefining2021}. However, skin texture can appear rough and bumpy when facial pores enlarge. There are various exogenous and endogenous factors, such as gender, race, aging, and hormones that cause enlarged facial pores. Higher casual sebum levels in the nose and medial cheek area explain larger pores observed from these areas \cite{leeFacial2016, uhodaConundrum2005}. Skincare products for pore care are widely available. Although there is a number of skincare diagnostic and recommendation capabilities available, to our knowledge, there is a lack of effective simulations today that reflect true facial pore transformation using available skincare products. Therefore, this paper proposes a complete pipeline to simulate the facial pore changes across a short-term period. Temporal analyses of changes in facial pores and simulation of these changes are helpful for the consumers to dynamically understand their skin and evaluate the potential benefit of skincare products. Such a realistic simulation not only guides customers to buy the right skincare products but also helps develop effective skincare technologies. Additionally, facial pore simulation can play a role, for example, in simulating the elongated effect of facial pores during the aging process. It enriches aging signs and contributes to the development of better aging models.

A realistic short-term simulation of the efficacy of skincare products builds upon truthful clinical studies. Similar to \cite{miyamotoDaily2021}, clinical studies were conducted on 60 young Japanese females to reveal  facial pore changes after applying specific skincare products A and B (in product code). The eMR Pro devices, designed for portable self-facial imaging with constant positioning and brightness \cite{miyamotoDaily2021}, were provided for participants to capture side-face images three times a day (morning after wake up; morning after face wash; and evening after face wash). Facial pore changes correlate with the initial skin condition, which varies among the 60 participants. Following \cite{DissanayakeNew2019}, various indexes are used to evaluate facial pore condition and we observe large perturbations in the index values due to daily skin fluctuations \cite{miyamotoDaily2021}. Based on this complex clinical data, we expect first to select valuable index(es) that reflect facial pore changes before and after using skincare products, regardless of different initial skin conditions. We further propose a sliding window mechanism to reduce the data perturbations efficiently and get the data ready for training.

\begin{figure}[!t]
\centering
    \includegraphics[trim=0cm 0cm 0cm 0cm,clip,width=0.8\columnwidth]{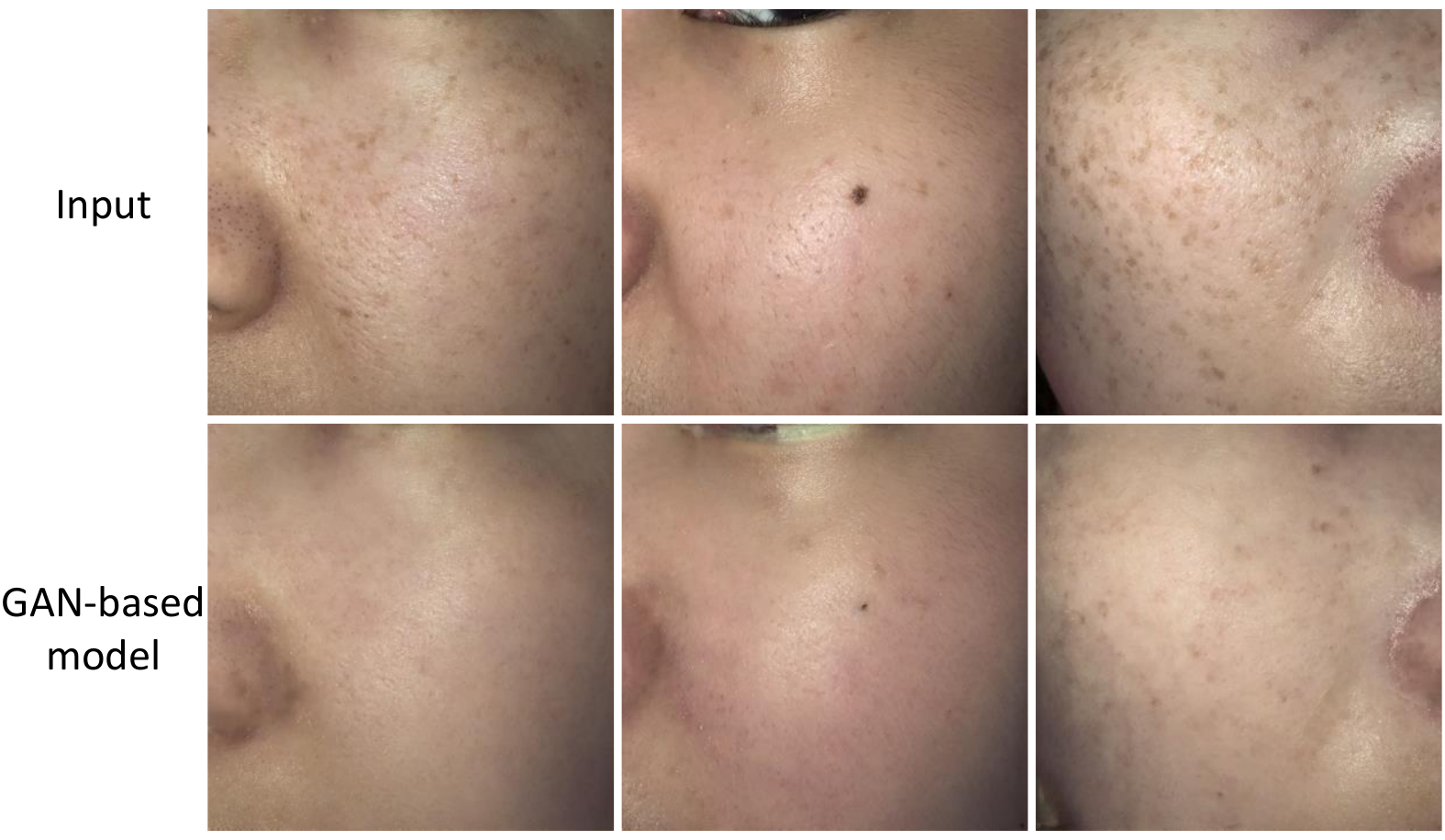}
  \caption{Sample results using a GAN-based method.}
  \vspace{-0.3cm}
  \label{Figure:gan}
\end{figure}

Our next step is to perform a facial pore simulation. To realistically simulate changes in facial pores while maintaining image fidelity, we need to pay attention to three aspects: morphological modification of all facial pores, precise control of the modification to reflect the real changes, and the non-pore areas remaining unchanged. Nowadays, GAN-based architectures \cite{KarrasStyleBased2019, KarrasAnalyzing2020, KarrasAliasFree2021} are famous for generating high-quality images. However, these methods have limitations in maintaining non-pore facial region features. And the facial pore transformations after using skincare products correlate with the initial condition, which is also challenging for GAN-based methods to capture such changes. From Figure~\ref{Figure:gan}, we can see that GAN-based architecture is weak at simulating the detailed variation of facial pores (blurring effect in the cheek area) and maintaining the non-pore facial features.

To address the above-mentioned challenges, we propose a facial pore simulation model that consists of three modules: segmentation module, regression module, and deformation module. The segmentation module is to provide accurate spatial information by detecting and localizing visible and enlarged facial pores. Existing pore detection works \cite{DissanayakeNew2019, WangImagebased2019, FrancoisQuantification2009, Jangmethod2018, SunAutomatic2017, ZhangSkin2008} take traditional approaches: setting threshold values carefully to segment facial pore features. Clinical imaging devices, such as, Visia-CR \cite{DissanayakeNew2019, Jangmethod2018} and Dermascore \cite{FrancoisQuantification2009} are used to capture high-resolution images at the same time to control lighting conditions and posture. \cite{WangImagebased2019, FrancoisQuantification2009, Jangmethod2018, SunAutomatic2017, ZhangSkin2008} are also limited to processing only small skin regions instead of full faces due to the threshold mechanism. By contrast, our UNet-based segmentation model is trained to learn robust facial pore features and to detect visible and enlarged pores in full side-face images. Our data-driven model detects pores with different sizes \& shapes, and it can handle images with different lighting to a certain extent. In parallel, a random forest regression model is expected to learn the pattern of facial pore changes over time from the training data and to further predict the time-dependent warping hyperparameter. The warping hyperparameter is to control the degree of facial pore deformation. Lastly, in the deformation module, the local scaling warp method \cite{GustafssonInteractive} is modified to calculate a flow-field grid with taking pore segmentation labels and warping hyperparameters as inputs. The flow-field grid assures that only facial pores are deformed, leaving the non-pore facial region unchanged. The original input image is then deformed accordingly to simulate facial pore changes after using specific skincare products. 

To summarize, the contributions of this work are as follows:
\vspace{-0.3cm}
\begin{itemize}
\item[$\bullet$]We propose a suite of customized analytical tools to process complex facial pore clinical data. We demonstrate that the sliding window mechanism helps clean the data by reducing perturbations due to significant skin fluctuations.
\item[$\bullet$]We propose a facial pore simulation that consists of three modules: segmentation module, regression module, and deformation module. We show that by incorporating facial pore segmentation labels and predicted warping hyperparameters, the deformation module can precisely deform facial pores with accurate control, while leaving the rest of the face unchanged. 
\item[$\bullet$]We propose a complete simulation pipeline that has two steps: training data establishment and facial pore simulation. Our results show that the simulation produces high-quality images which demonstrate realistic facial pore changes over time.
\end{itemize}



\section{Related Work}
\subsection{Facial Pore detection}
Facial pore detection is a challenging task as facial pores are delicate with different shapes (circular or elongated) \cite{Jangmethod2018, Shaieknew2017} \& sizes (from 50 $\upmu$m to 500 $\upmu$m) \cite{CarlosUsing2014}. It requires images to be sharp enough to contain pore-level features. Existing works \cite{DissanayakeNew2019, FrancoisQuantification2009, Jangmethod2018, SunAutomatic2017, ZhangSkin2008, KimSebum2013, Shaieknew2017, Flamentautomatic2019} require professional equipment to capture high-quality facial skin in a fixed pose under consistent light.
\cite{FlamentArtificial2022, Flamentautomatic2019} only output the overall grading of facial pores. Traditional approaches are mainly used to detect facial pores: mark pores on melanin layers derived from digital images \cite{CarlosUsing2014, WangFacial2019}; using the difference of Gaussian (DoG) filters \cite{DissanayakeNew2019}; based on Fast fuzzy c-mean algorithm \cite{ZhangSkin2008}. However, these methods are constrained to work on skin patches, and often perform poorly on images with varied lighting. Our proposed UNet-based model is capable of detecting skin pores on full side-face images with satisfactory performance and addresses different lighting problems.

\subsection{Face Simulation}
Existing GAN-based works \cite{ShiCANGAN2020, WuFace2022, MakhmudkhujaevReAging2021, LiuAttributeAware2019, YangLearning2018, FangTripleGAN2020} focus on simulating the human face aging process. For instance, facial attributes are used to guide wavelet-based GANs \cite{LiuAttributeAware2019} to simulate aging effect. \cite{YangLearning2018} learns about people's age progression by unraveling subject-specific features and age-specific effects. Triple-GAN \cite{FangTripleGAN2020} proposes to learn the interrelationships between different age groups. Aging simulation is at a larger scale than simulating skin changes. To the best of our knowledge, no work has been done to simulate the efficacy of skincare products on the face. 

\begin{figure*}[!t]
\centering
  \begin{minipage}{0.6\textwidth}
  \includegraphics[trim=2cm 1.1cm 2.5cm 1.1cm,clip, width=0.75\columnwidth]{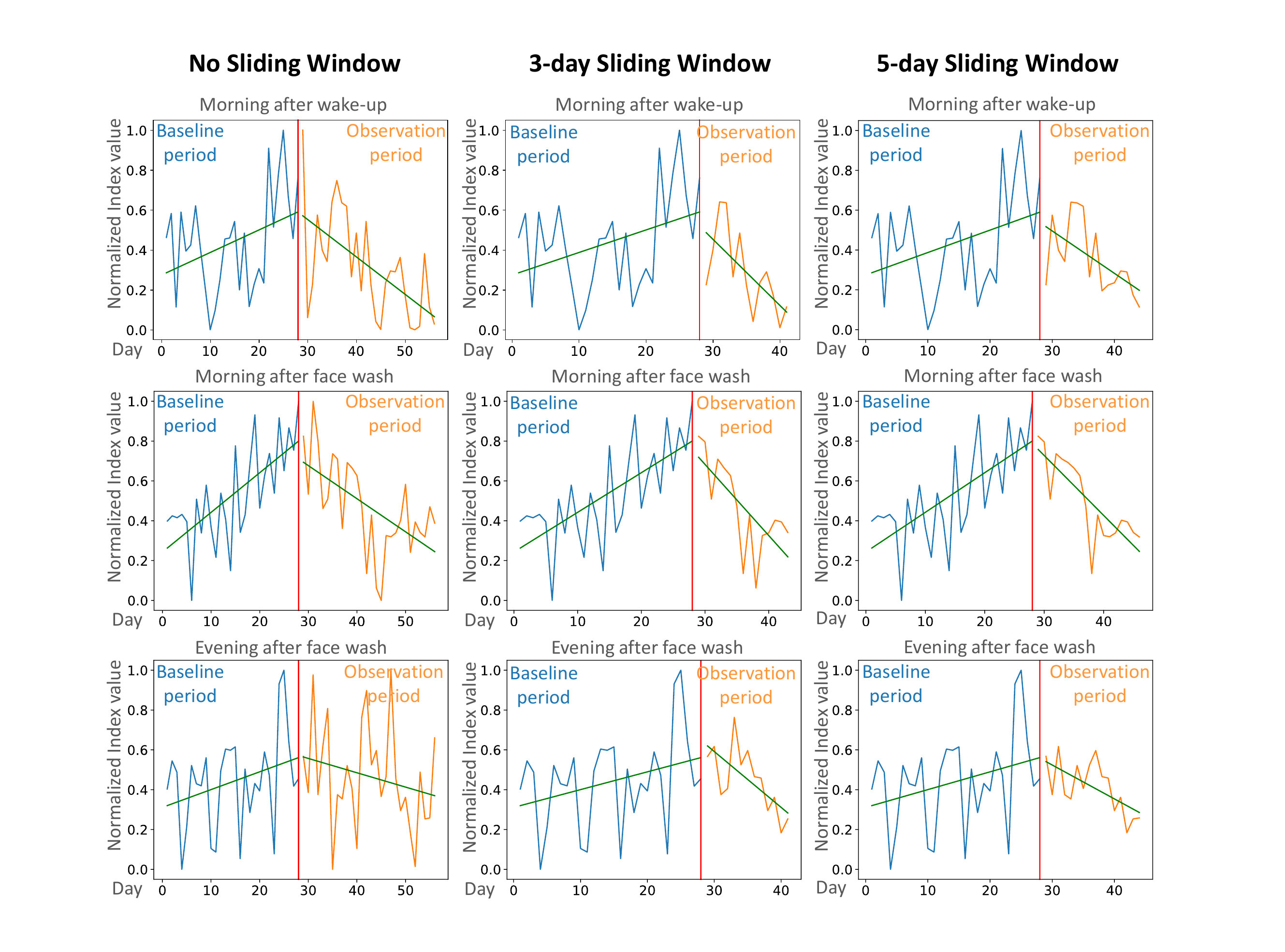}
  \raisebox{0.4\height}{\includegraphics[width=0.2\columnwidth]{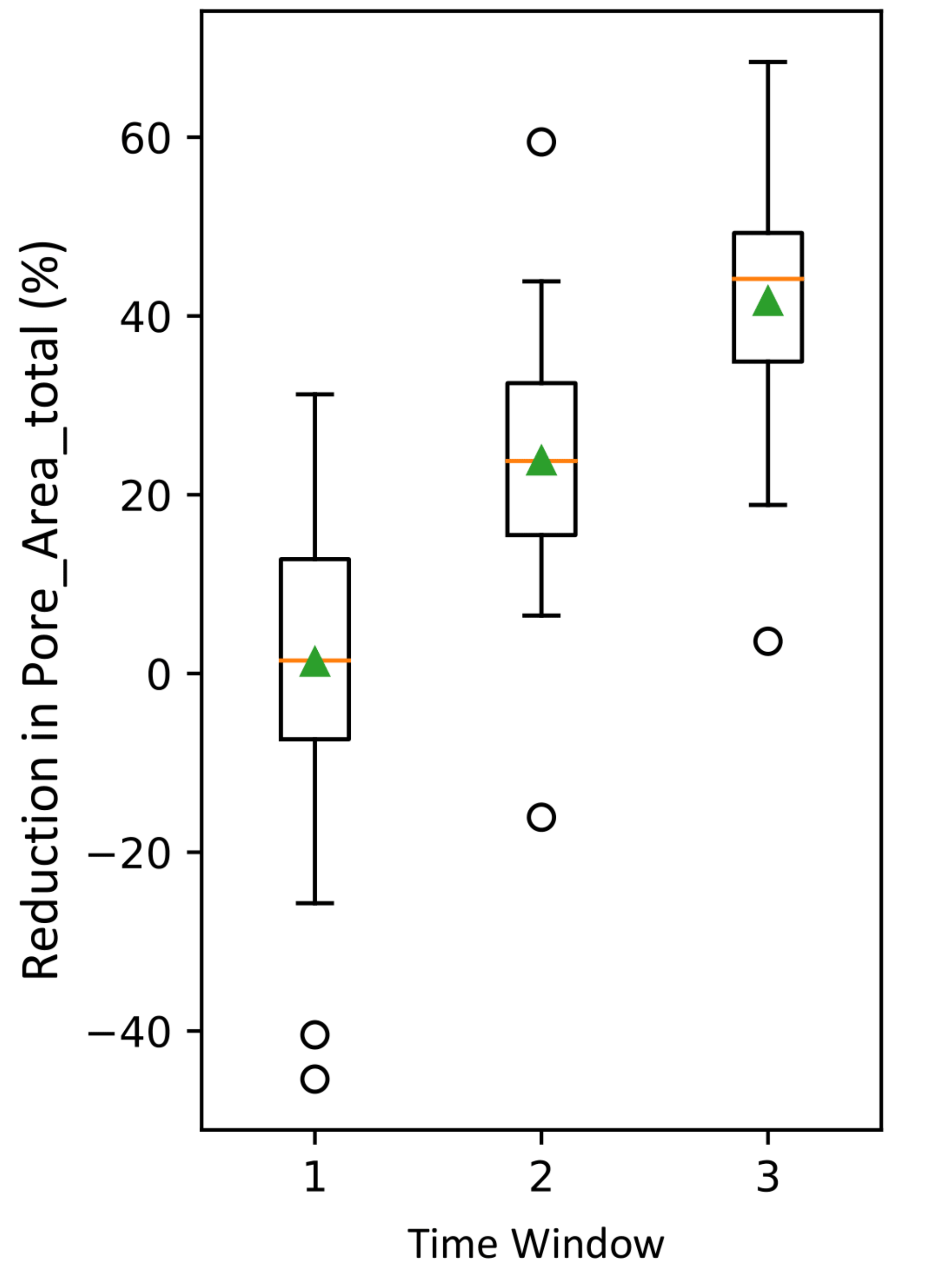}}
  \captionsetup{justification=centering}
  \vspace{0.1cm}
  \caption{Pore\_Area\_total index analysis.}
  \label{Figure:stats}
  \end{minipage}
\begin{minipage}{0.35\textwidth}
{
\includegraphics[trim=0.5cm 0.1cm 0cm 0.1cm,clip,width=1\columnwidth]{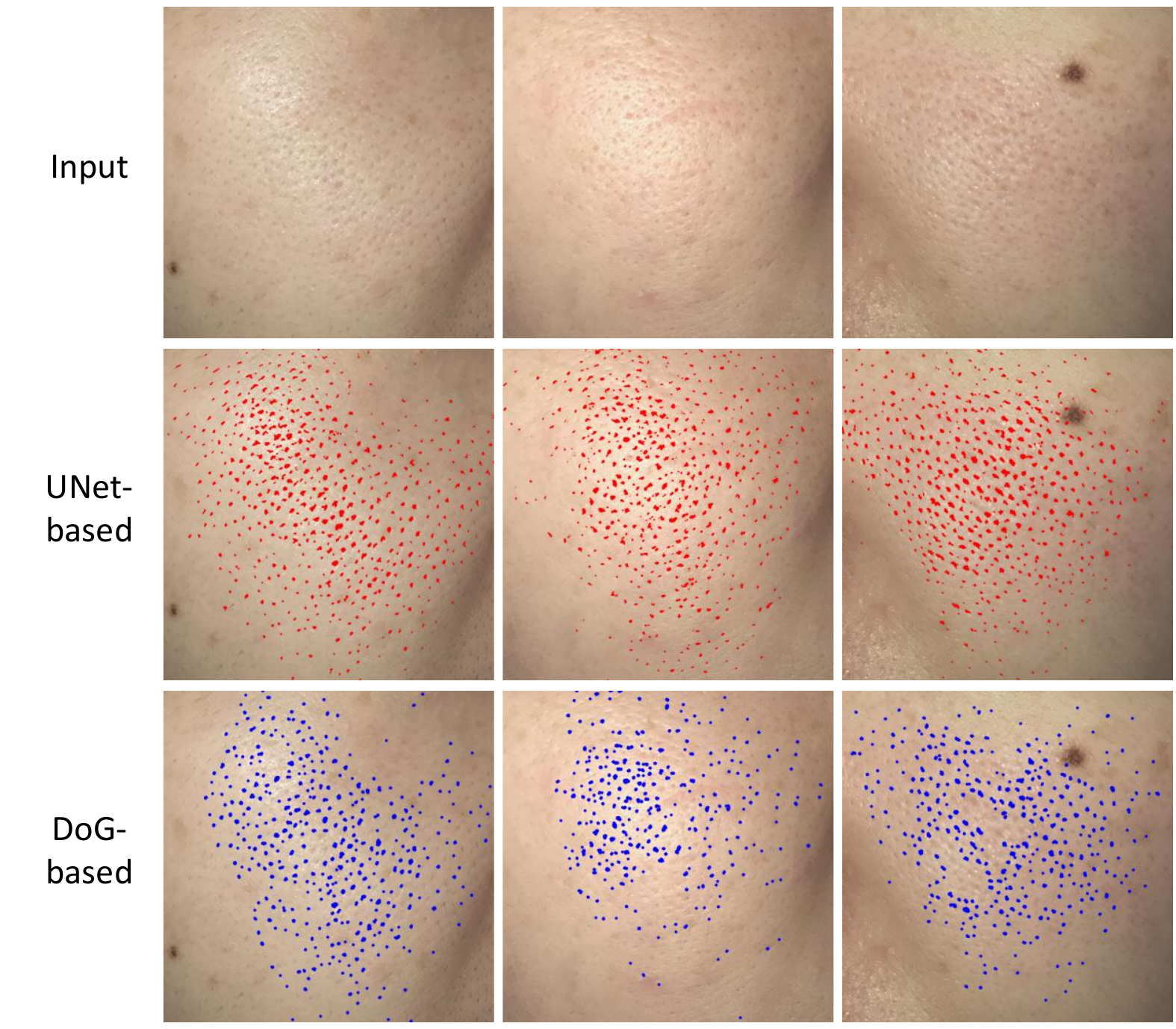}}
  \captionsetup{justification=centering}
  \vspace{0.2cm}
  \caption{Qualitative results of Facial Pore Segmentation models.}
  \label{Figure:segCmp}
  \end{minipage}
  \vspace{-0.3cm}
\end{figure*}

\section{Methodology}
\subsection{Training Data Establishment}
8-week clinical studies have been conducted on 60 young Japanese females from 22 to 34 years old: they used product A and/or B on the left/right side of their faces. Participants used eMR Pro devices \cite{miyamotoDaily2021} connected to smartphones to capture side-face images three times a day: morning after wake-up; morning after face wash; evening after face wash. Various pore statistics were measured using the method from \cite{DissanayakeNew2019}: pore count; total pore area; mean pore area; pore shape; orientation; mean L*/a*/b*-channel value in CIELAB color space. \cite{miyamotoDaily2021} examines the noticeable fluctuations in the skin that people experience every day, especially with facial pores. We use similar approaches from \cite{miyamotoDaily2021} to analyze these measured indexes while focusing on the 4-week observation period with applying products A and B. Our dataset consists of 12,531 images in total, and participants have different initial skin conditions with diverse pore size scales (from grade 0 to grade 5) \cite{Shaieknew2017}. To observe the facial pore changes among participants equally, we first normalize index value for every participant. On each observation day, a mean value of each index is computed to represent facial pore condition among the group. This averaging operation is designed to consider the different initial skin conditions of 60 participants. Linear regression analysis is then conducted to study index patterns before and after using skincare products. Valuable index(es) are to be selected to represent facial pore changes. Considering people have different degrees of skin fluctuation across days, we propose a $n$-day sliding window mechanism to reduce the data perturbations: gradually group $n$-day values in 1-day steps and then remove extreme values by the $n$-day mean and standard deviation. In addition, we observe that facial pore changes are developed slowly over the 4-week period of using skincare products, and we propose to split the 4-week observational duration into 3 discrete time windows to investigate skincare efficacy. The cleaned dataset contains 3,025 images and each participant has multiple sets of images to reflect their facial pore changes across time windows, which is used as the training set to develop a facial pore simulation model.

\subsection{Facial Pore Simulation}
\subsubsection{UNet-based Facial Pore Segmentation}
Pore segmentation labels lay the foundation for developing a good simulation of facial pore changes. In contrast with existing threshold-based traditional works \cite{DissanayakeNew2019, WangImagebased2019, FrancoisQuantification2009, Jangmethod2018, SunAutomatic2017, ZhangSkin2008}, we propose a deep learning-based data-driven model to detect visible and enlarged facial pores on side-face images captured by smartphones. We use 9,506 out of 12,531 images as the segmentation training set. It is time-consuming and labor-intensive if we were to manually annotate facial pores. Additionally, the simulation does not require exhaustive detection of facial pores. Hence, we propose a new approach to generate satisfactory labels to improve productivity. Inspired by \cite{ChenDeep2020}, we use the Photocopy filter to extract facial details and then apply various post-processing operations to reduce excessive noise: morphological dilation and erosion; setting thresholds for pore area, size \cite{CarlosUsing2014}, and shape \cite{Jangmethod2018, Shaieknew2017}. Lastly, the pore segmentation labels were manually checked and the necessary refinements were made.

Witnessing the outstanding performance of UNet-based architectures in medical image segmentation, we train a pore segmentation model on 9,506 images using UNet \cite{RonnebergerUNet2015a} architecture. By measuring the similarity between the predicted segmentation label and the true segmentation label, the loss function penalizes the model to learn in the optimal direction. We add cross entropy loss and F1 score to form our loss function:
\begin{equation}
\label{eq:loss}
\small
\textrm{$\mathscr{L}$} = \textrm{Cross Entropy} + \textrm{F1 Score}.
\end{equation}
Our selfie images are captured at 1920 x 1080 pixels. To avoid large memory overhead over training, images are cropped into patches with size of 256 x 256 with batch size of 24. RMSProp optimization algorithm is used with learning rate = 1e-5, weight decay = 1e-8, and momentum = 0.9. 

\subsubsection{Random Forest Regression}
The regression module is to build up the relationship between the selected index and time window statistically. Our data indicate that participants experience different facial pore transformations with varied initial skin conditions, making it challenging to build an appropriate regression model. In addition, our data are considerably large in size. Therefore, we exploit Random Forest \cite{BreimanRandom2001} as our regression model. Random Forest consists of a forest of classifying decision trees: it selects a subset of data randomly over training at each split and efficiently improves accuracy and controls over-fitting by using average values based on all aggregations \cite{Fernandez-Delgadowe, Speisercomparison2019}.

\begin{figure*}[!t]
\centering
  \includegraphics[width=1\textwidth]{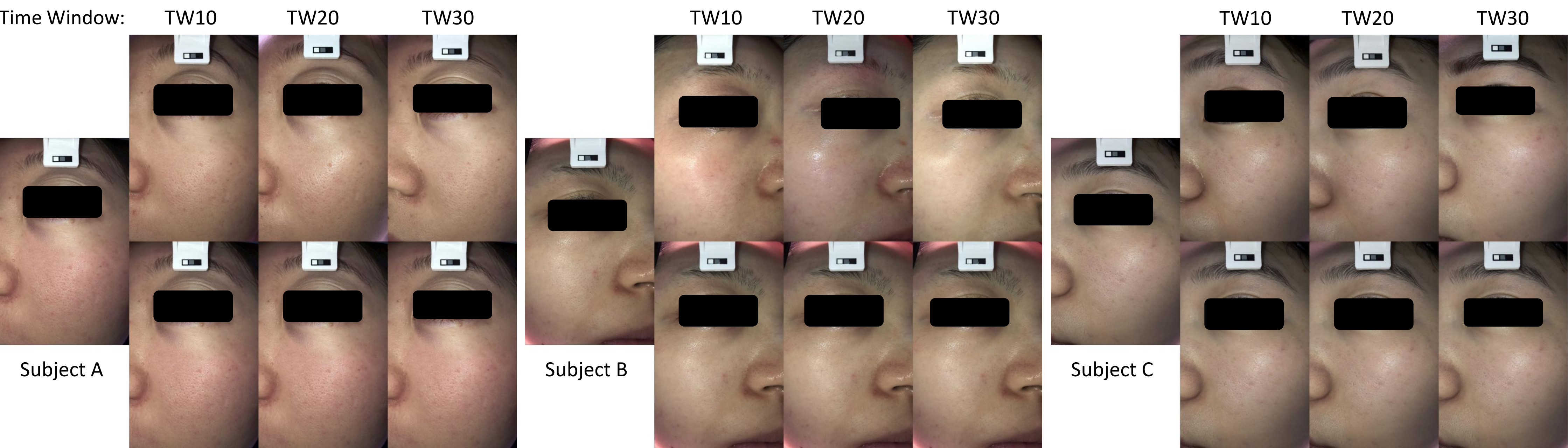}
  \caption{Qualitative result of Facial Pore Simulation.}
  \label{Figure:simu_results}
\end{figure*}

\subsubsection{Facial Pore Deformation} 
Local scaling warp from \cite{GustafssonInteractive} creates the possibility for precise manipulation. It achieves pixel-wisely manipulation by mapping points from input space to warping space without changing the colors, and the mapping is computed by:

\begin{equation}
\label{eq:warp_eq}
\small
f_{s}(r)=\left(1-\left(\frac{r}{r_{\max }}-1\right)^{2} a\right) r.
\end{equation}
One minimum enclosing circle is found for each detected pore and the radius of each enclosing circle is $r_{max}$. Variable $r$ measures the distance between the target pixel and the center of the enclosing circle, and parameter $a$ controls the degree of deformation. Instead of implementing this method in an interactive manner, a well-fit random forest regression model is trained to predict parameter $a$. Based on Equation~\ref{eq:warp_eq}, mappings for all detected pores are gathered to compute a flow-field grid. Each size-2 vector in the flow-field grid is used to interpolate the corresponding output value. The bilinear interpolation method is used here and border values are used for out-of-bound grid locations. The face deformation operation is also optimized and it takes only about 5-10 seconds to manipulate one full side-face image of size 1920 x 1080.


\section{Results}
Our complete simulation pipeline consists of 2 stages: training data establishment and facial pore simulation. First, we show the analytical patterns of facial skin quality indexes over time. Next, we present the evaluation for facial pore segmentation, random regression, and facial pore simulation.

\subsection{Training Data Establishment}
Various indexes were measured in digital images. People's skin fluctuates a lot within a day and/or from day to day, and we can observe large data perturbations in pore quality index values. After careful analysis of the statistics, we observe that index \emph{Pore\_Area\_total} demonstrates a jittery decreasing pattern across 3 time windows (shown in the first column in Figure~\ref{Figure:stats}), and this pattern is consistent with the visual analysis by domain experts. Index \emph{Pore\_Area\_total} is then selected as the representative index to reflect facial pore changes after skincare. With implementing the sliding window mechanism in the observation period, outliers are removed, and the data display fewer perturbations (shown in the second and third columns of Figure~\ref{Figure:stats}). The 3-day sliding window mechanism retains most data and it is then utilized to clean the data. After splitting the 4 observation week into 3 time windows, we further investigate the changes in value of index Pore\_Area\_total. The box plot also shows a jittery decreasing pattern in Figure~\ref{Figure:stats}, and it supports our observation. 

\begin{table}[!h]
\caption{Qualitative Evaluation in Face Pore Segmentation}
\vspace{-0.3cm}
\label{tab:seg results}
\small
\begin{center}       
\begin{tabular}{p{0.2\columnwidth}<{\centering}|p{0.25\columnwidth}<{\centering}|p{0.25\columnwidth}<{\centering}} 
\hline
Metrics & UNet-based & DoG-based \\ \hline
Dice &  \textbf{0.6663}  & 0.4418 \\ \hline
IoU &  \textbf{0.5139}  & 0.2851 \\ \hline
Precision &  \textbf{0.6790}  & 0.4388 \\ \hline
Accuracy &  \textbf{0.9936}  & 0.9896 \\ \hline
\end{tabular}
\vspace{-0.75cm}
\end{center}
\end{table}

\subsection{Facial Pore Simulation}
\subsubsection{UNet-based Facial Pore Segmentation}
A test set of 30 images is manually annotated for quantitative evaluation of facial pore segmentation. These test images are carefully selected from the cleaned facial pore simulation training set (consisting of 3,025 images), which considers different facial pore conditions and lighting conditions. As existing works are implemented in traditional threshold-based methods, we only compare our UNet-based pore segmentation model with the DoG-based method \cite{DissanayakeNew2019}. Multiple standard metric results in Table~\ref{tab:seg results} show that the UNet-based segmentation model outperforms the DoG-based methods by a large margin. Figure~\ref{Figure:segCmp} shows the visual comparison and we can conclude that our UNet-based segmentation model is capable of segmenting facial pores of different sizes and shapes. The model is also performing well in facial cheek area where we can observe specular effect in the first two input images in Figure~\ref{Figure:segCmp}.

\subsubsection{Random Forest Regression}
Random forest regression is trained to learn the relationship between metric \emph{Pore\_Area\_total} and time windows. $R^2$ Score and Mean Average Error (MAE) are utilized here to evaluate regression performance. The results in Table~\ref{tab:regression} show the random forest regression model fits well with low error. 

\begin{table}[!h]
\vspace{0.2cm}
\caption{Random Forest Regression Analysis}
\vspace{-0.3cm}
\small
\label{tab:regression}
\begin{center}   
\begin{tabular}{m{0.125\columnwidth}<{\centering}|m{0.22\columnwidth}<{\centering}|m{0.22\columnwidth}<{\centering}|m{0.22\columnwidth}<{\centering}} 
\hline
Metrics & Time Window 1 & Time Window 2 & Time Window 3 \\ \hline
$R^2$ Score & 0.9905 & 0.9941 & 0.9564 \\ \hline
MAE & 0.6344±2.20 & 0.6039±1.85 & 1.3498±5.12 \\ \hline
\end{tabular}
\end{center}
\end{table} 

\begin{table}[!h]
\caption{Quantitative Evaluation on Simulation Image Quality}
\vspace{-0.2cm}
\label{tab:simulation quality}
\small
\begin{center}       
\begin{tabular}{p{0.25\columnwidth}<{\centering}|p{0.4\columnwidth}<{\centering}} 
\hline
Metrics & Simulation images \\ \hline
PIQE & 11.86 \\ \hline
NIQE & 3.41 \\ \hline
\end{tabular}
\vspace{-0.8cm}
\end{center}
\end{table}

\subsubsection{Facial Pore Deformation}
As we have discussed previously, existing face simulation methods focus on age regression and their evaluation metrics are not applicable here. Similar to image super-resolution \cite{ZhangRankSRGAN, MengMagFace2021}, face image quality is a significant factor to develop a high-quality simulation system. In our dataset, side-face images are inevitably misaligned even with using the eMR Pro devices \cite{miyamotoDaily2021}. In addition, image registration methods do not work well for side-face images. Therefore, we use No-Reference Image Quality Assessment (NR-IQA) for evaluation: NIQE \cite{MittalMaking2013} and PIQE \cite{NBlind2015}. The detailed results in Table~\ref{tab:simulation quality} indicate the high quality of the simulated images. In Figure~\ref{Figure:simu_results}, we show qualitative results of our model simulating the facial pore changes of 3 subjects after using skincare products. For each subject, the leftmost image shows the condition of his skin before using skin care products. Real images shown in the top row demonstrate his real skin condition after 10/20/30 days of skincare. Correspondingly, the bottom row presents the simulated skin condition of the facial pores for each time window (TW10, TW20, TW30). As can be seen, our model captures the delicate facial pore changes across the three time windows and modifies the input image to reflect realistically his facial pore changes over time. We can also observe that the model only modifies facial pore appearance while leaving other facial features unchanged, as shown in Figure~\ref{Figure:simu_results}.

\section{Perception Study}

\begin{figure*}[!t]
\centering
  \begin{minipage}{0.31\textwidth}
  \includegraphics[trim=2.6cm 0.2cm 2cm 0.2cm,clip, width=1\columnwidth]{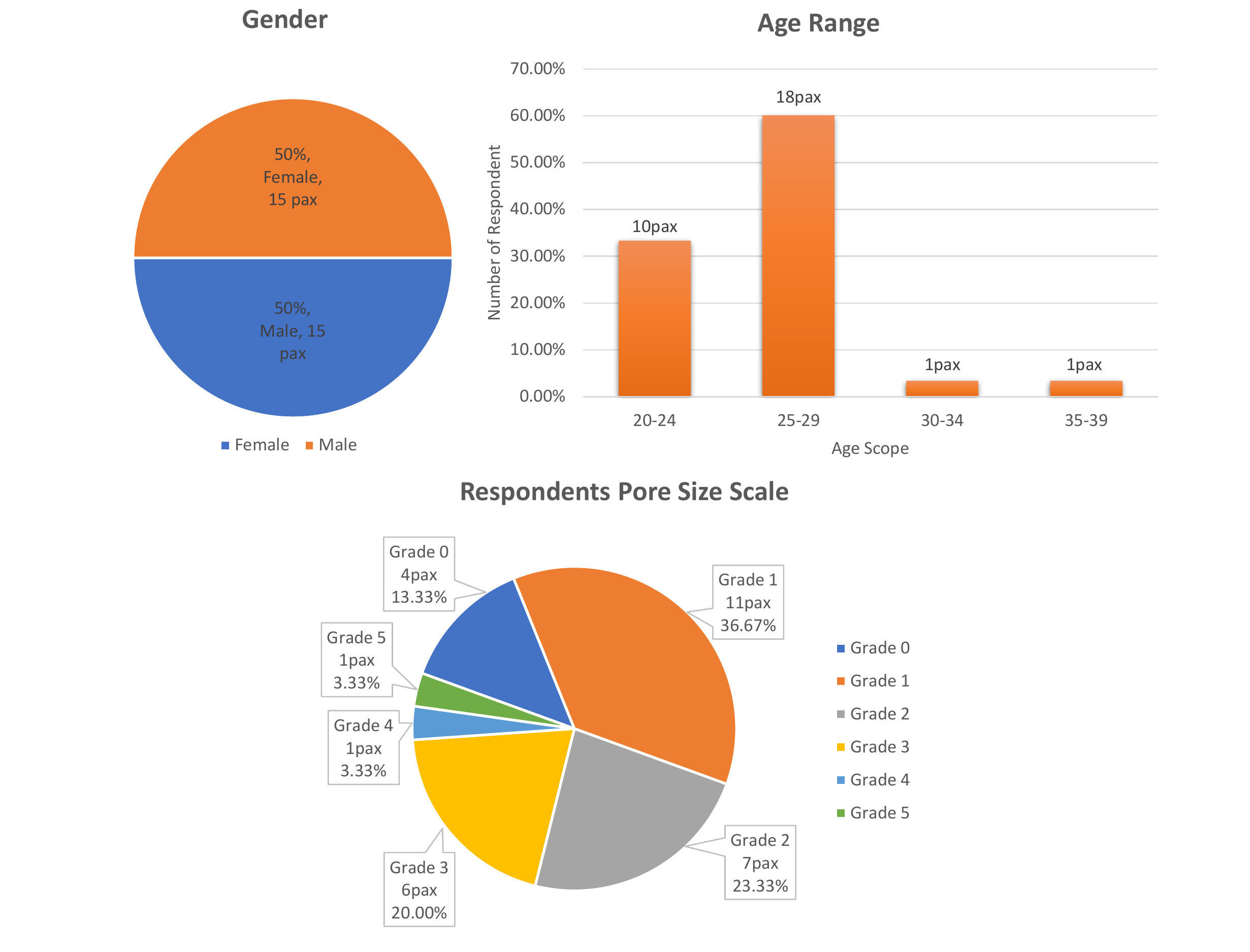}
  \captionsetup{justification=centering}
  \caption{Meta information of respondents.}
  \label{Figure:meta}
  \end{minipage}
\begin{minipage}{0.31\textwidth}
  \includegraphics[trim=2.6cm 0.2cm 2cm 0.2cm,clip, width=1\columnwidth]{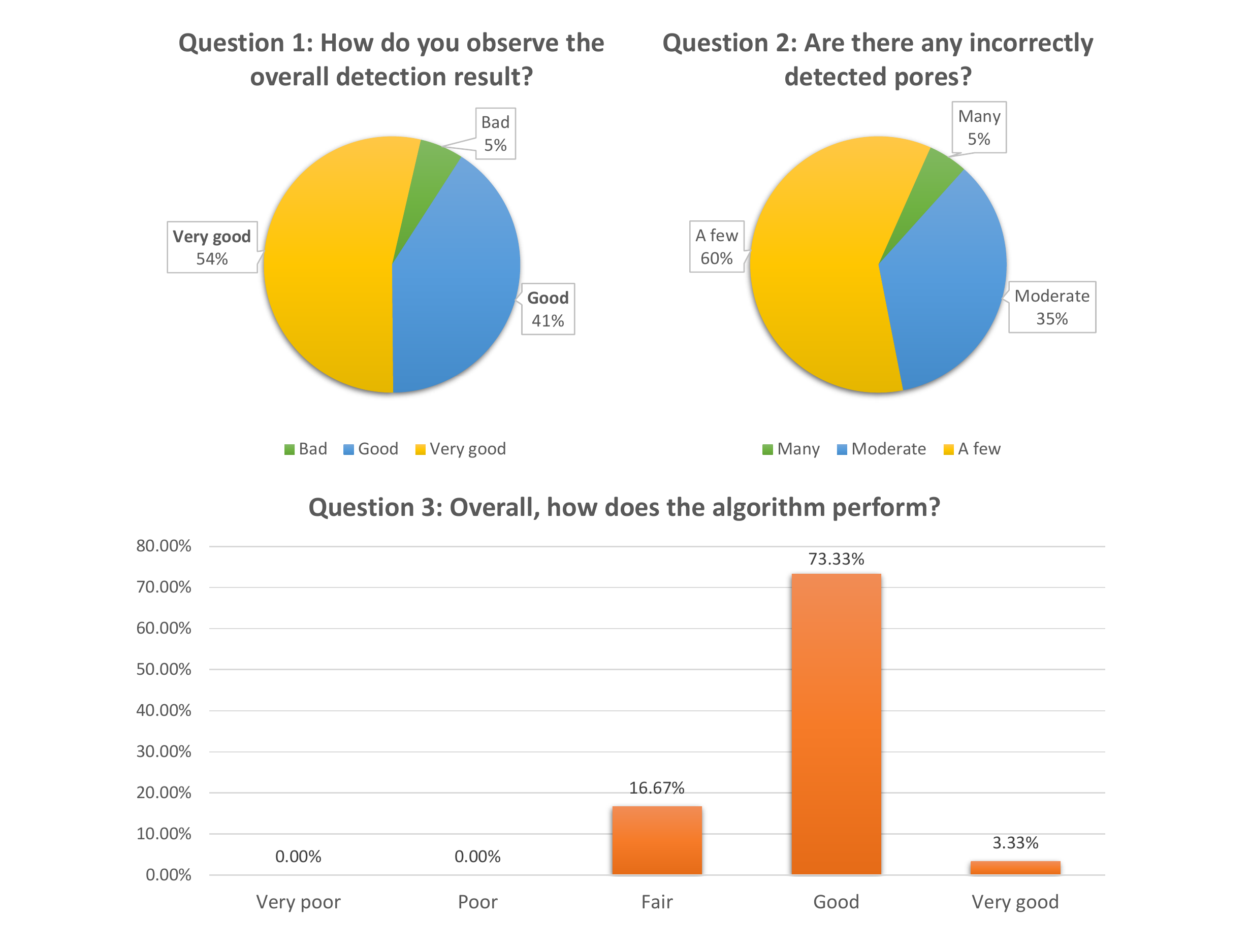}
  \captionsetup{justification=centering}
  \caption{Respondents' Score on Facial Pore Segmentation.}
  \label{Figure:seg}
  \end{minipage}
\begin{minipage}{0.31\textwidth}
  \includegraphics[trim=2.6cm 0.2cm 2cm 0.2cm,clip, width=1\columnwidth]{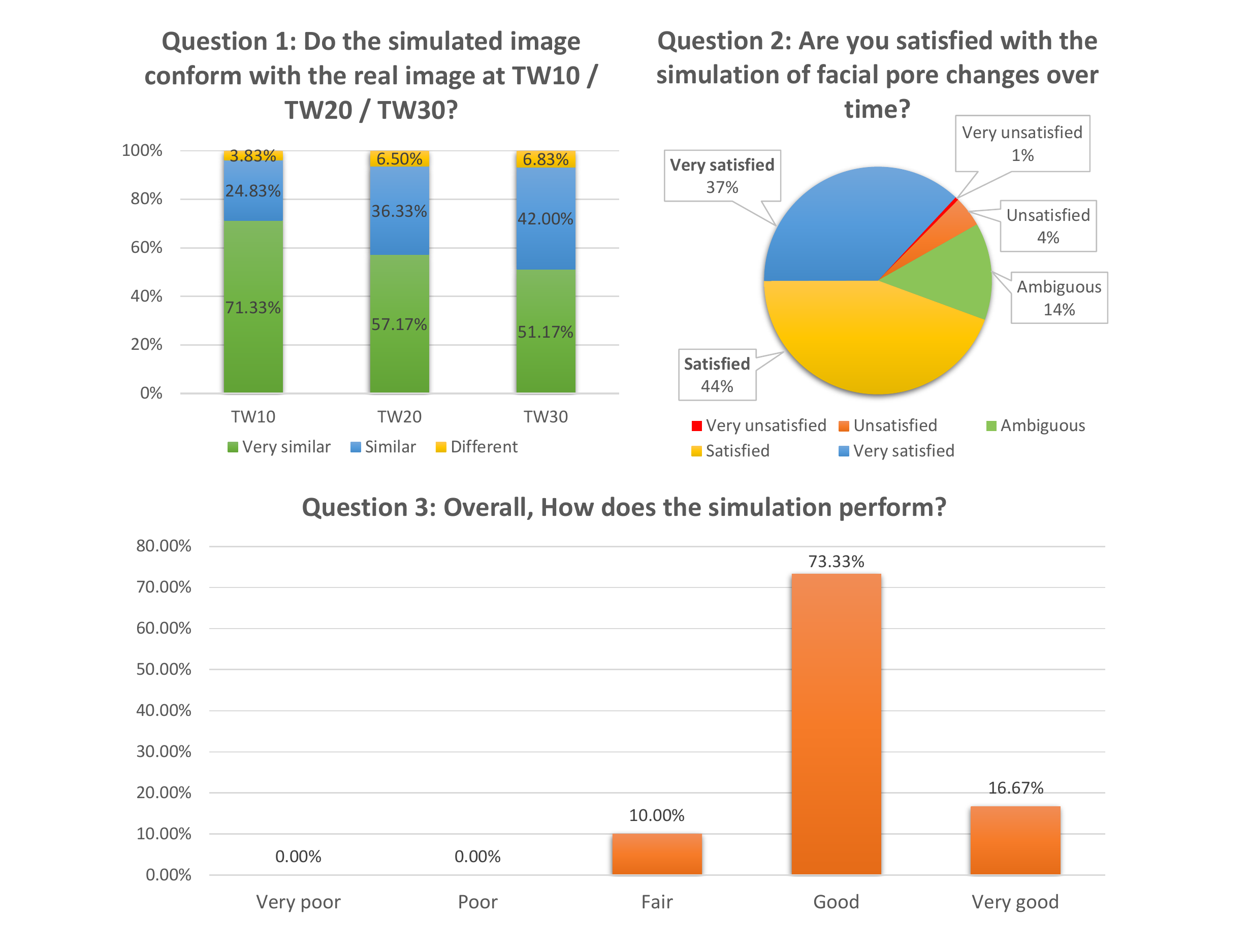}
  \captionsetup{justification=centering}
  \caption{Respondents' Score on Facial Pore Simulation.}
  \label{Figure:simu}
  \end{minipage}
  \vspace{-0.2cm}
\end{figure*}

We further conduct a perception study to evaluate our UNet-based segmentation model and facial pore simulation model. To be fair, we selected 40 images of facial pore segmentation results with diverse pore conditions under different light conditions. Similarly, 20 sets of simulation images were prepared. 

We invited 30 respondents to this study: 15 women and 15 men, from ethnicity: Chinese and Indian. From the age range graph in Figure~\ref{Figure:meta}, we can see that more than 90\% of the respondents are in their 20s, which is consistent with the young Japanese participants in our clinical studies. In addition, we asked the respondents to carefully self-assess their facial pore status with reference to a clinical standardized scale of pore size from \cite{Shaieknew2017}. The pore size scale pie chart in Figure~\ref{Figure:meta} shows that the facial pore status is diverse among the 30 respondents, which shares a similar distribution to our clinical study.

\subsection{Facial Pore Segmentation}
Respondents were to ask 3 questions when observing 40 pore segmentation results:
\begin{itemize}
\item[$\smallbullet$]Q1: How do you observe the overall detection results? (Choose one: Very Good; Good; Bad)
\item[$\smallbullet$]Q2: Are there any incorrectly detected pores? (Choose one: A few; Moderate; Many)
\item[$\smallbullet$]Q3: Overall, how does the algorithm perform? (Choose one: Very Good; Good; Fair; Poor; Very Poor)
\end{itemize}

The first and second questions are to evaluate the performance of the UNet-based pore segmentation model in terms of false negative and false positive respectively. Respondents answered Q1 and Q2 for each image and grade the overall performance in Q3 after viewing 40 images. As shown in Figure~\ref{Figure:seg}, 95\% of the responses show satisfaction with our pore detection results (54\% Very Good; 41\% Good; and 5\% Bad). In Q2, 60\% of responses indicate there are a few false positives in segmentation results and 35\% indicate a moderate number of false positives. Overall, more than 75\% of the respondents feel good or very good at facial pore segmentation, with no negative comments.

\subsection{Facial Pore Simulation}
Our simulation progressively predicts facial pore changes at each time window. Respondents were invited to answer 3 questions by seeing 20 sets of images to evaluate each time window as well as the entire process.

\begin{itemize}
\item[$\smallbullet$]Q1: Do the simulated images conform with the real images at TW10 (\textit{i.e.,} Time Window of 10-day using skincare)/TW20/TW30? (Choose one: Very Similar; Similar; Different)
\item[$\smallbullet$]Q2: Are you satisfied with the simulation of facial pore changes over time? (Choose one: Very Satisfied; Satisfied; Ambiguous; Unsatisfied; Very unsatisfied)
\item[$\smallbullet$]Q3: Overall, how does the simulation perform? (Choose one: Very Good; Good; Fair; Poor; Very Poor)
\end{itemize}

\vspace{-0.4cm}

In Q1, respondents compared real and simulated images at each time window and scored their similarity in terms of facial pore condition only. And Q2 focused on whether the simulation model captured the pattern of facial pore changes across time. Respondents then graded the overall performance after viewing 20 sets of images in Q3. Figure~\ref{Figure:simu} shows that more than 90\% of the responses vote that the simulated images are similar or very similar to the real images at each time window. Over 80\% of respondents rate the simulated facial pore changes as satisfied or very satisfied by looking at the 1-month duration. The bottom plot from Figure~\ref{Figure:simu} shows that 90\% of the respondents (27 persons) perceive that this simulation faithfully reflected the facial pore changes over 4 weeks of skincare use.

\section{Conclusion}
In this paper, we present a realistic short-term facial pore simulation model to evaluate the efficacy of skincare products. We demonstrate that the sliding window mechanism is useful in reducing data perturbations caused by daily fluctuations in skin. The complex facial pore clinical data is then reorganized and prepared for training. Moreover, to achieve a realistic simulation on facial pores while maintaining image fidelity (\textit{i.e.,} the non-pore facial areas remain unchanged), we propose a facial pore simulation that precisely localizes visible and enlarged facial pores and further modify facial pores morphologically to reflect facial pore changes after applying skincare product over a short-term. Our method delivers promising results for facial pore segmentation and facial pore simulation.

\section{Acknowledgements}
This work was carried out at the Rapid-Rich Object Search (ROSE) Lab, Nanyang Technological University (NTU), Singapore. The research is supported in part by A*STAR under it’s A*STAR MBRC Strategic Positioning Fund (SPF) – A*STAR-P\&G Collaboration (Award APG2013/113) and in part by China-Singapore International Joint Research Institute under Grant 206-A018001. Any opinions, findings and conclusions or recommendations expressed in this material are those of the author(s) and do not reflect the views of the A*STAR. 





\end{document}